\documentclass[runningheads]{llncs}

\usepackage{eccv}

\usepackage{eccvabbrv}

\usepackage{graphicx}
\usepackage{booktabs}

\usepackage{multirow}
\usepackage{wrapfig}
\usepackage{comment}

\usepackage[accsupp]{axessibility}  

\usepackage[normalem]{ulem}
\useunder{\uline}{\ul}{}
\newcommand{\mycustomsize}{\fontsize{6pt}{7pt}\selectfont}

\definecolor{dgreen}{HTML}{0aa344}
\definecolor{dblue}{HTML}{4b5cc4}
\definecolor{dred}{HTML}{be002f}

\usepackage{hyperref}

\usepackage{orcidlink}

\begin{document}

\title{Faceptor: A Generalist Model for Face Perception} 



\author{Lixiong Qin\orcidlink{0009-0003-3899-6245} \and
Mei Wang\orcidlink{0000-0002-3559-9346} \and
Xuannan Liu\orcidlink{0009-0005-1428-3261} \and
Yuhang Zhang\orcidlink{0000-0003-4161-5020} \and
Wei Deng\orcidlink{0009-0004-3886-6907} \and
Xiaoshuai Song\orcidlink{0009-0003-8870-1759} \and
Weiran Xu\orcidlink{0000-0002-9416-7666}\thanks{Corresponding author.} \and
Weihong Deng\orcidlink{0000-0001-5952-6996}
}

\authorrunning{L. Qin et al.}

\institute{Beijing University of Posts and Telecommunications, Beijing, China
\email{\{lxqin,wangmei1,xuweiran,whdeng\}@bupt.edu.cn}}

\maketitle

\begin{abstract}
With the comprehensive research conducted on various face analysis tasks, there is a growing interest among researchers to develop a unified approach to face perception.
Existing methods mainly discuss unified representation and training, which lack task extensibility and application efficiency. 
To tackle this issue, we focus on the unified model structure, exploring a face generalist model.
As an intuitive design, \textbf{Naive Faceptor} enables tasks with the same output shape and granularity to share the structural design of the standardized output head, achieving improved task extensibility.
Furthermore, \textbf{Faceptor} is proposed to adopt a well-designed single-encoder dual-decoder architecture, allowing task-specific queries to represent new-coming semantics. 
This design enhances the unification of model structure while improving application efficiency in terms of storage overhead.
Additionally, we introduce Layer-Attention into Faceptor, enabling the model to adaptively select features from optimal layers to perform the desired tasks. 
Through joint training on 13 face perception datasets, Faceptor achieves exceptional performance in facial landmark localization, face parsing, age estimation, expression recognition, binary attribute classification, and face recognition, achieving or surpassing specialized methods in most tasks.
Our training framework can also be applied to auxiliary supervised learning, significantly improving performance in data-sparse tasks such as age estimation and expression recognition. 
The code and models will be made publicly available at \url{https://github.com/lxq1000/Faceptor}.
\end{abstract}

\section{Introduction}
\label{sec:intro}

In recent years, substantial strides have been made in face perception research.
Numerous methods have been developed to enhance performance in face analysis tasks such as facial landmark localization~\cite{SLPT,DTLD}, face parsing~\cite{AGRNet,DML-CSR}, age estimation~\cite{MWR,DLDLv2}, expression recognition~\cite{KTN,EAC}, binary attribute classification~\cite{MCNN-AUX,DMM-CNN} and face recognition~\cite{SphereFace,ArcFace,CosFace}.
There are several concerns related to these methods which necessitate a distinct deep model for each task.
Firstly, from a methodological perspective, it is not cost-effective to conduct large-scale data collection and model training for each face analysis task due to the fact that there is only one object of interest - the human face.
Secondly, from a practical perspective, real-world applications often simultaneously require a set of face analysis tasks to cater to specific businesses. It is inefficient to deploy numerous models.

\begin{figure}[tb]
    \centering
    \begin{subfigure}{0.45\textwidth}
        \centering
        \includegraphics[height=3cm]{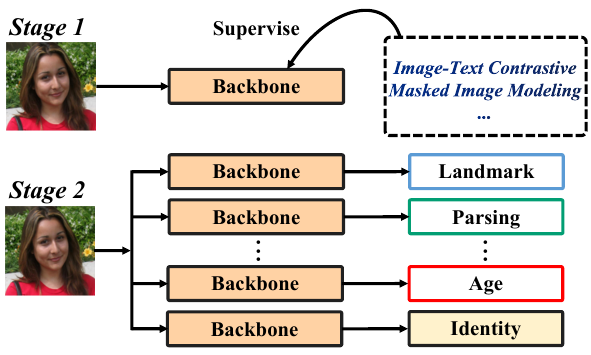}
        \caption{Unified representation: \\universal representation + finetuning\centering}
        \label{fig:main1-a}
    \end{subfigure}
    \begin{subfigure}{0.45\textwidth}
        \centering
        \includegraphics[height=3cm]{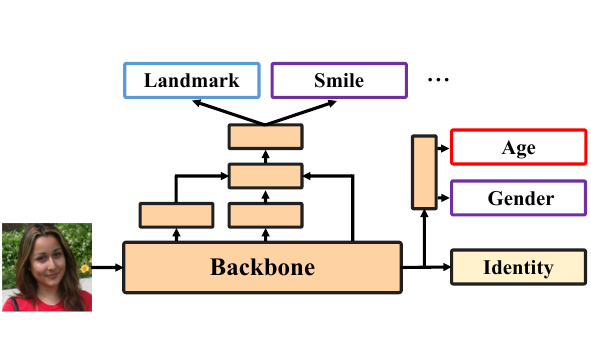}
        \caption{Unified training: \\early all-in-one model + multi-task learning\centering}
        \label{fig:main1-b}
    \end{subfigure}
    \begin{subfigure}{0.45\textwidth}
        \centering
        \includegraphics[height=3cm]{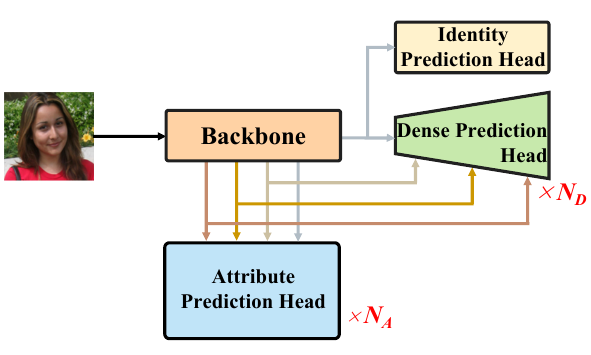}
        \caption{Unified model structure (ours, shared structural designs): Naive Faceptor\centering}
        \label{fig:main1-c}
    \end{subfigure}
    \begin{subfigure}{0.45\textwidth}
        \centering
        \includegraphics[height=3cm]{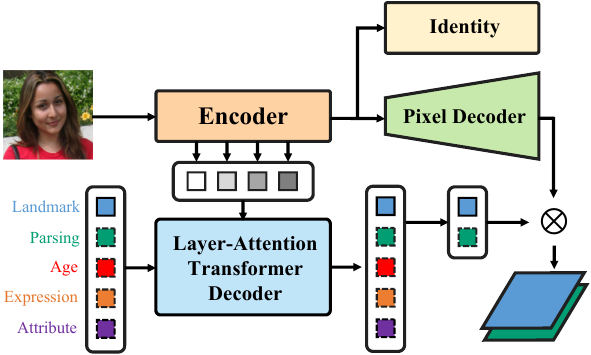}
        \caption{Unified model structure (ours, shared parameters): Faceptor\centering}
        \label{fig:main1-d}
    \end{subfigure}
    \caption{Existing efforts for unified face perception mainly concentrate on representation and training. Our work focuses on unified model structure, achieving improved task extensibility and increased application efficiency by two designs of face generalist models.}
    \label{fig:main1}
\vspace{-15pt}
\end{figure}

In light of this, researchers have naturally turned their attention toward achieving a unified approach for face perception.
Existing efforts mainly concentrate on the following two aspects:
(1) Unified representation.
As shown in \cref{fig:main1-a}, FRL~\cite{FRL} and FaRL~\cite{FaRL} initially obtain a task-agnostic backbone through universal facial representation learning (unsupervised learning~\cite{SwAV}, self-supervised learning~\cite{Beit,MIM}, and natural language supervised learning~\cite{CLIP,ALIGN,NLS}).
By avoiding the need to collect large-scale datasets specifically for supervised pre-training of each task, these approaches improve data efficiency. 
However, they still require separate finetuning for each downstream task, resulting in low application efficiency in terms of the training process, inference speed, and storage overhead.
(2) Unified training. 
As shown in \cref{fig:main1-b}, HyperFace~\cite{HyperFace} and AIO~\cite{AIO} employ a multi-task learning framework to simultaneously handle a predefined set of face analysis tasks, eliminating the repetitiveness in model training.
However, due to the empirically determined output structures for each task, these early all-in-one models are unable to address new-coming tasks, resulting in a lack of task extensibility. 
Furthermore, these early models lack robust pre-training and are now considered to have performed inadequately.

In this work, we aim to explore a face generalist model, which is initialized with a task-agnostic backbone (unified representation) and can handle any user-chosen set of face analysis tasks with a multi-task learning framework (unified training).
To achieve improved task extensibility and increased application efficiency, we laser-focus on the unified model structure.
Two ideas are presented as follows:

(1) Shared structural designs: dealing with new-coming tasks using standardized output heads.
We have observed significant variations in the expected outputs of different face analysis tasks in terms of shape and granularity.
Based on these observations, we categorize all face analysis tasks into three distinct categories: dense prediction, attribute prediction, and identity prediction. 
An intuitive model design can consist of a backbone and three types of standardized output heads, each dedicated to a specific task category, as illustrated in \cref{fig:main1-c}, referred to as \textbf{Naive Faceptor}.
All tasks share a common backbone, enabling the proposed model to achieve higher application efficiency than the unified representation approaches.
Tasks within the same category will share structural designs, thus avoiding the need to design new output structures based on experience for new-coming tasks, and ensuring the extensibility of the model. 
However, a notable limitation of this design is the lack of parameter sharing among heads across tasks. This results in a linear growth of the number of heads as the tasks increase, leading to significant storage overhead.

(2) Shared parameters: dealing with new-coming semantics using task-specific queries.
To further enhance the unification of model structure while maintaining the model's performance on individual tasks, we propose \textbf{Faceptor}, which adopts a single-encoder dual-decoder architecture, as shown in \cref{fig:main1-d}.
The transformer encoder extracts shared features while the transformer decoder attends to particular semantic information.
Additionally, the pixel decoder is used for restoring the image spatial scale for dense prediction tasks.
Inspired by previous works~\cite{DETR,MaskFormer,Mask2Former,SLPT,Rethinking}, we introduce task-specific queries from single-task methods into our unified structure to model the semantics of different tasks, minimizing the use of non-shared parameters and achieving a significantly higher storage efficiency.
We also introduce the Layer-Attention mechanism in the transformer decoder to model the preferences of different tasks towards features from different layers.
With layer-aware embeddings introduced into the transformer decoder, Faceptor can adaptively assign weights for the features from different layers.

In multi-task learning, the objective is to achieve optimal performance across all tasks, while auxiliary supervised learning leverages some tasks to enhance the performance of others.
In our training framework, auxiliary supervised learning can be performed by adjusting the weights and batch sizes of involved tasks.
Our experimental findings indicate that harnessing facial landmark localization, face parsing and face recognition tasks can significantly enhance the performance of tasks such as age estimation and expression recognition, which suffer from limited available data.

Our contributions can be summarized as follows:
\begin{enumerate}
    \item  To the best of our knowledge, our work is the first to explore a face generalist model, with unified representation, training, and model structure.
    Our main focus is on the development of unified model structures.
    \item With one shared backbone and three types of standardized output heads, \textbf{Naive Faceptor} achieves improved task extensibility and increased application efficiency.
    \item With task-specific queries to deal with new-coming semantics, \textbf{Faceptor} further enhances the unification of model structure and employs significantly fewer parameters than Naive Faceptor. 
    \item The proposed Faceptor demonstrates outstanding performance under both multi-task learning and auxiliary supervised learning settings.
  \end{enumerate}

\section{Related Works}
\label{sec:related_works}

\textbf{Universal Facial Representation: }
FRL~\cite{FRL} and FaRL~\cite{FaRL} address face analysis tasks by following a pipeline that involves (1) collecting a large-scale facial dataset, (2) pre-training a task-agnostic network to achieve universal facial representation learning, and (3) fine-tuning the network for specific facial tasks in the user-chosen set.
FRL~\cite{FRL} adopts the unsupervised learning method SwAV~\cite{SwAV}, which simultaneously clusters the data while enforcing consistency between the cluster assignments produced for different augmentations of the same image.
FaRL~\cite{FaRL} combines natural language supervised and self-supervised learning, extracting high-level semantic meaning from image-text pairs using contrastive loss~\cite{CLIP,ALIGN,NLS}, while also exploring low-level information through masked image modeling~\cite{Beit,MIM}. 
Robust pre-training is crucial for face generalist models.
In our experiments, we utilize the ViT~\cite{ViT} model pre-trained with the FaRL framework as the initialization for the transformer encoder.

\textbf{Multi-task Learning for Face Perception: }
Multi-task learning is initially analyzed in detail by Caruana~\cite{Multitask}.
In the field of face perception, HyperFace~\cite{HyperFace} and AIO~\cite{AIO} are early classic works of multi-task learning, employing CNN as the backbone and leveraging experiential knowledge to determine the appropriate layer of features for different tasks.
However, since these models are designed for predefined task sets, they are not able to deal with new-coming tasks.
In contrast, SwinFace~\cite{SwinFace} adopts standardized subnets for task extensibility, with face analysis and recognition subnets handling attribute and identity prediction tasks respectively.
In our experiments, the Naive Faceptor is primarily inspired by SwinFace but includes an additional subnet~\cite{UperNet} to handle dense prediction tasks.

\textbf{Transformer Encoder-Decoder Architecture for Computer Vision: }
The success of DETR~\cite{DETR} in object detection has motivated researchers to investigate the utilization of transformer encoder-decoder architecture in computer vision tasks.
MaskFormer~\cite{MaskFormer} presents a unified approach to tackle semantic and instance-level segmentation tasks through the introduction of a single-encoder dual-decoder structure. In MaskFormer, each segment is represented by a query in the transformer decoder.
In SLPT~\cite{SLPT} and RLPFER~\cite{Rethinking}, individual facial landmarks or expressions are considered distinct semantic information and are represented as task-specific queries.
To the best of our knowledge, there is no existing work in the field of face perception that comprehensively unifies all face analysis tasks and employs task-specific queries to represent diverse semantic information.

\section{Method}
\label{sec:method}
In this section, we first offer a brief introduction to the structure of Naive Faceptor.
Next, we provide the details of the Faceptor design, highlighting the Layer-Attention mechanism.
Then, we present the training framework and discuss the objective functions.
Lastly, we provide a comprehensive comparison between our proposed face generalist models and previous efforts for face perception.

\subsection{Naive Faceptor}
\label{sec:naive_faceptor}
We briefly describe the structure of Naive Faceptor. 
For a fair comparison, the backbone of Naive Faceptor and the encoder of Faceptor utilize the same transformer encoder architecture, initialized by the FaRL~\cite{FaRL} framework.
Details regarding the transformer encoder will be provided in \cref{sec:faceptor}.
We employ standardized face analysis and face recognition subnets from SwinFace~\cite{SwinFace} as attribute prediction head and identity prediction head, respectively.
In addition, we follow the implementation in the FaRL experiment, utilizing UperNet~\cite{UperNet} as the dense prediction head to produce dense output.
We provide an illustration of Naive Faceptor in the appendix, offering more details.

\subsection{Faceptor}
\label{sec:faceptor}
\begin{figure}[tb]
\centering
\includegraphics[height=6.0cm]{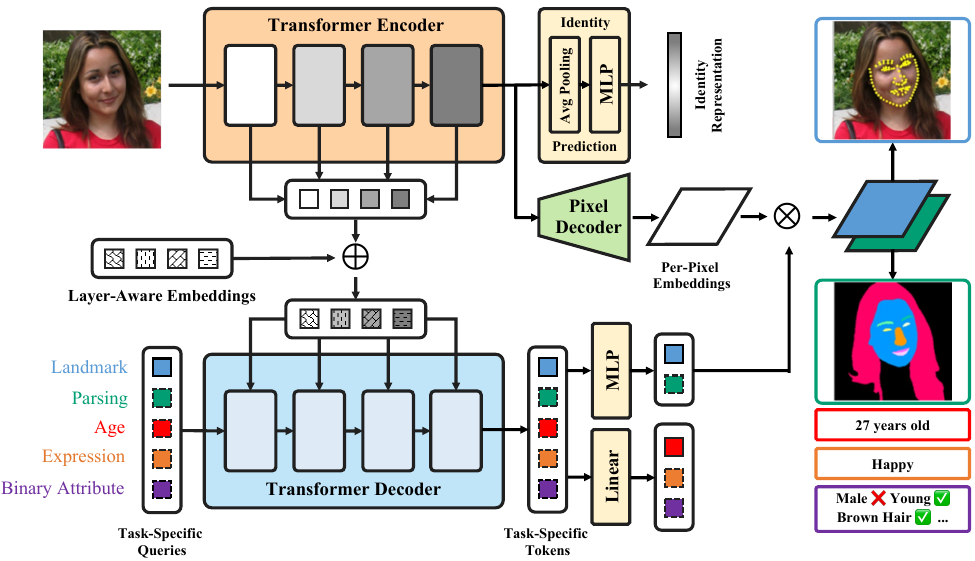}
\caption{Overall architecture for the proposed Faceptor}
\label{fig:main2}
\vspace{-15pt}
\end{figure}

Faceptor adopts a single-encoder dual-decoder architecture, as shown in \cref{fig:main2}.

\subsubsection{Transformer Encoder}
We utilize a 12-layer ViT-B~\cite{ViT} as the transformer encoder, which is pre-trained with FaRL~\cite{FaRL} framework.
When an image $\mathbf{X}$ of size $H \times W$ is given as input, the encoder produces a feature $\mathbf{F}^l\in \mathbb{R} ^{C_{en}\times \frac{H}{S} \times \frac{W}{S}}$ at the $l$-th layer. 
Here, $C_{en}$ represents the number of channels, and $S$ represents the stride of patch projection, which are 768 and 16 respectively.
To handle input images of varying resolutions ($512\times512$ for dense prediction tasks, and $112\times112$ for attribute and identity prediction tasks), we employ a shared learnable positional embedding $\mathbf{E}_{en_-pos}$ with a size of $32\times32$, and interpolate it based on the spatial size of the input image after patch projection.
We retain the features obtained from all 12 layers of the encoder for future use. 
Therefore, the encoded feature F can be formulated as: 
\begin{equation}
    \mathbf{F}=\mathrm{TransformerEncoder}(\mathbf{X},\mathbf{E}_{en\_pos})\in \mathbb{R}^{12\times C_{en}\times\frac{H}{S}\times \frac{W}{S}},
    \label{eq:encoder}
\end{equation}
where $\mathbf{F}=[\mathbf{F}^1;\mathbf{F}^2;\cdots;\mathbf{F}^{12}]$.

\subsubsection{Transformer Decoder}
We employ a 9-layer standard transformer decoder~\cite{Attention} to compute the task-specific tokens based on the encoded features and task-specific queries. 
To begin, we define task-specific queries, which are applicable to dense prediction and attribute prediction tasks. The task queries for task $t$ are denoted as:
\begin{equation}
    \mathbf{Q}_t=[\mathbf{q}_{t,1},\mathbf{q}_{t,2},\mathbf{q}_{t,3},...,\mathbf{q}_{t, N_t}],
    \label{eq:queries}
\end{equation}
where $N_t$ represents the number of queries that convey different semantic meanings in task $t$. 
A landmark, a semantic parsing class, and a binary attribute are each represented by one query for facial landmark localization, face parsing, and binary attribute classification respectively.
101 queries represent ages 0-100 for age estimation. 
7 queries represent expressions (surprise, fear, disgust, happiness, sadness, anger, neutral) for expression recognition.  
Following established conventions~\cite{Attention,Mask2Former}, all task-specific queries $\mathbf{Q}_t$ are accompanied by a positional embedding $\mathbf{E}_{de\_pos,t}$, which has the same dimension as $\mathbf{Q}_t$ and is not shared across tasks.

Typically, when using the transformer decoder in visual tasks, only the encoded feature from the top layer, denoted as $\mathbf{F}^{top}$, is utilized for computation. 
However, the features obtained from the encoder contain decreasing geometric information and increasing semantic information from the bottom to the top layers. Different tasks have varying preferences for features from different layers.
To enable the transformer decoder to leverage features from multiple layers, we uniformly extract six layers of features from $\mathbf{F}$ and project them into the dimension of the decoder tokens, denoted as $C_{de}$ and set to 256, resulting in:
\begin{equation}
    \hat{\mathbf{F}}=\mathrm{Projection}([\mathbf{F}^2;\mathbf{F}^4;\mathbf{F}^6;\mathbf{F}^8;\mathbf{F}^{10};\mathbf{F}^{12}])\in \mathbb{R}^{6\times C_{de}\times\frac{H}{S}\times \frac{W}{S}}.
    \label{eq:f_hat}
\end{equation}
After processing with the transformer decoder, task-specific tokens for dense prediction or attribute prediction task $t$ are obtained:
\begin{equation}
    \mathbf{T}_t=\mathrm{TransformerDecoder}(\hat{\mathbf{F}},\mathbf{Q}_t,\mathbf{L}_t,\mathbf{P},\mathbf{E}_{de\_pos,t})\in \mathbb{R}^{N_t\times C_{de}},
    \label{eq:transformer_decoder}
\end{equation}
where $\mathbf{L}_t$ and $\mathbf{P}$ are the layer-aware embedding and positional embedding associated with $\hat{\mathbf{F}}$, respectively. 
Further details are provided in \cref{sec:layer_attention}.

\subsubsection{Pixel Decoder}
The pixel decoder is used to gradually upsample the features in order to produce per-pixel embeddings:
\begin{equation}
\mathbf{E}_{pixel}=\mathrm{PixelDecoder}(\mathbf{F})\in \mathbb{R}^{C_{de}\times \frac{H}{s}\times \frac{W}{s}},
    \label{eq:pixel_decoder}
\end{equation}
where $s$ is set to 4 in our implementation.
It should be noted that any per-pixel classification-based segmentation model can be employed as a pixel decoder.
In our implementation, we extract the feature $\mathbf{F}^{12}$ from the top layer of the encoder, and then pass it through two consecutive $2\times2$ deconvolutional layers to obtain the per-pixel embedding $\mathbf{E}_{pixel}$.
Experimental results have demonstrated that this simple pixel decoder has been capable of achieving excellent performance in facial landmark localization and face parsing.

\subsubsection{Outputs}
Similar to Naive Faceptor, Faceptor also includes specifically designed output modules for three categories of tasks. 
For the dense prediction tasks, the task-specific tokens need to be passed through a shared MLP to align with the per-pixel embeddings outputted by the pixel decoder. 
The dot product of these two is then linearly interpolated to obtain the final dense prediction output $\mathbf{y}_{map} \in \mathbb{R}^{N_t \times H \times W}$.
For the attribute prediction tasks, the task-specific tokens produced by the decoder can directly go through a shared linear layer to obtain the final prediction result $\mathbf{y}_{value} \in \mathbb{R}^{N_t}$.
For the identity prediction task, the features from the top layer of the transformer, denoted as $\mathbf{F}^{12}$, are first passed through an average pooling layer to obtain a vector. 
Then, following the implementation of SwinFace~\cite{SwinFace}, the vector is processed by an FC-BN-FC-BN structure to obtain the final identity representation $\mathbf{y}_{vector}\in\mathbb{R}^{d}$, where $d$ is set to 512.
It is important to note that in Faceptor, all parameters of output modules are shared among multiple tasks of the same category, whereas in Naive Faceptor, tasks of the same category share only the structural design of output modules without sharing parameters.

\subsection{Layer-Attention Mechanism}
\label{sec:layer_attention}

In the transformer decoder, cross-attention can be represented as: 
\begin{equation}
    \mathrm{CrossAttention}(\mathbf{Q},\mathbf{K},\mathbf{V})=\mathrm{Softmax}(\mathbf{Q}\mathbf{K}^T/\sqrt d)\mathbf{V}.
    \label{eq:cross_attention}
\end{equation}
For the $l$-th layer, the query is $\mathbf{Q}=\mathbf{H}_t^{l-1}+\mathbf{E}_{de\_pos,t}$,
where $\mathbf{H}_t^{l-1}$ is the output of the previous layer of the decoder and $\mathbf{H}_t^0=\mathbf{Q}_t$. 
The value is $\mathbf{V}=\hat{\mathbf{F}}$.
We implement Layer-Attention by introducing layer-aware embeddings $\mathbf{L}_t\in\mathbb{R}^{6\times C_{de}}$ for task $t$ into the key, obtaining:
\begin{equation}
    \mathbf{K}=\hat{\mathbf{F}}+\mathrm{Repeat}(\mathbf{L}_t)+\mathrm{Repeat}(\mathbf{P}),
    \label{eq:k}
\end{equation}
where $\mathbf{P} \in \mathbb{R}^{C_{de}\times\frac{H}{S} \times \frac{W}{S}}$ is the learnable positional embeddings randomly initialized, and the Repeat function extends the input features in a repeated manner to a scale of $\mathbb{R}^{6\times C_{de}\times\frac{H}{S} \times \frac{W}{S}}$.

For simplification, we use $\hat{\mathbf{L}}_t$ and $\hat{\mathbf{P}}$ to represent $\mathrm{Repeat}(\mathbf{L}_t)$ and $\mathrm{Repeat}(\mathbf{P})$ respectively.
In \cref{eq:cross_attention}, $\mathbf{Q}\mathbf{K}^T$ can be expanded as $\mathbf{Q}\hat{\mathbf{F}}^T+\mathbf{Q}\hat{\mathbf{L}}_t^T+\mathbf{Q}\hat{\mathbf{P}}^T$. 
The term $\mathbf{Q}\hat{\mathbf{P}}^T$ reflects the model's preference for features at different positions, typically taken into account by existing models.
In contrast, $\mathbf{Q}\hat{\mathbf{L}}^T_t$ represents the model's preference for features from different layers, which has often been neglected in previous research.

\begin{wrapfigure}{r}{0.6\textwidth}
  \centering
  \vspace{-15pt}
  \includegraphics[width=0.6\textwidth]{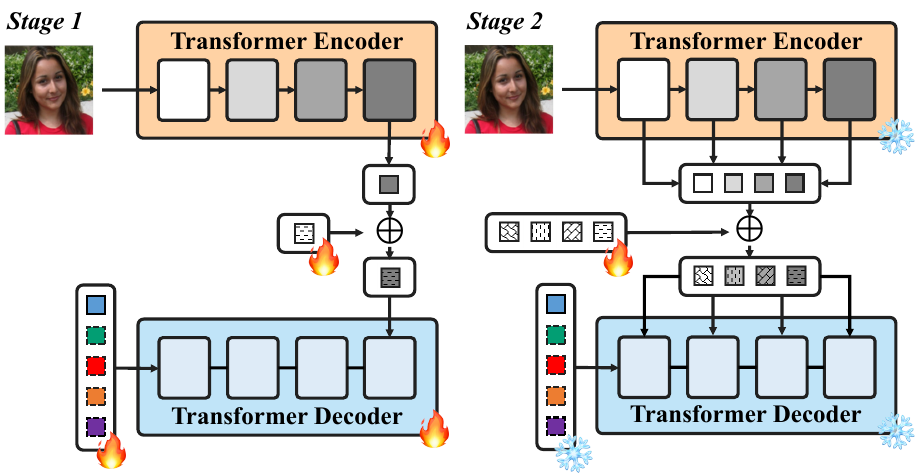}
  \caption{Two-stage training process to ensure the effectiveness of Layer-Attention mechanism.}
  \vspace{-15pt}
  \label{fig:main3}
\end{wrapfigure}
In practice, we found that directly introducing Layer-Attention can not improve the model's performance on various tasks, and even result in significant deterioration in the age estimation task. 
We believe that this is because both $\mathbf{Q}_t$ and $\mathbf{E}_{de\_pos}$ are randomly initialized, which causes, at the beginning of training, $\mathbf{Q}_t$ to be unable to represent semantic information and $\mathbf{Q}\hat{\mathbf{L}}_t^T$ to be inadequate in reflecting task $t$'s preference for features from different layers.
To address this issue, we introduce a two-stage training process, as shown in \cref{fig:main3}. 
In the first stage, only the features from the top layer, namely, $\mathrm{Projection}(\mathbf{F}^{12})$, are used for training to enable $\mathbf{Q}_t$ to learn the semantic representation of task $t$. 
In the second stage, the transformer decoder is allowed to access $\hat{\mathbf{F}}$, and most of the model parameters are frozen except for $\mathbf{L}_t$, which is allowed to be learned.
It should be noted that since $\mathbf{L}_t$ is not shared across tasks, if there is no performance improvement on task $t$ after the second stage of training, the Layer-Attention mechanism can be excluded during inference for task $t$.
Experimental results show that attribute prediction tasks such as age estimation, expression recognition, and binary attribute classification can benefit from the introduction of the Layer-Attention mechanism.

\subsection{Objective Functions}
We employ a multi-task learning framework to enable the model to simultaneously tackle a variety of face analysis tasks. The overall objective function is: 
\begin{equation}
    L_{all} = \frac{\sum_{t\in T}\alpha_t
    \frac{1}{n_t}\sum_{i=1}^{n_t}L(\mathbf{y}_{t,i})}{\sum_{t\in T}\alpha_t},
    \label{eq:overall_loss}
\end{equation}
where $T$ represents the user-chosen task set, 
$\alpha_t$ is the weight of task $t$, 
$n_i$ is the number of samples for task $t$ in each training batch,
$\mathbf{y}_{t,i}$ is the output of Faceptor for the $i$-th sample in task $t$,
and $L(\mathbf{y}_{t,i})$ is the loss function for single sample.
Auxiliary supervised learning can be performed by adjusting the $\alpha_t$ and $n_i$.
Please refer to the appendix for the specific loss function used for each individual task.

\subsection{Comparison of Task Extensibility and Application Efficiency}

\Cref{table:paradigm} presents a semi-quantitative comprehensive comparison between our proposed models and previous unified approaches in task extensibility and application efficiency.
Assuming there are $N$ tasks in the user-chosen set.
It is noticed that the number of parameters in the queries is much less than that in the output modules.
As $N$ increases, the number of parameters in Faceptor will be significantly less than that in Naive Faceptor.
To sum up, our Faceptor can achieve improved task extensibility and the highest application efficiency.

\vspace{-10pt}

\begin{table}[!h]\scriptsize
    \caption{Semi-quantitative comparison of task extensibility and application efficiency. \textcolor{dblue}{$\mathcal{B}$} represents backbones, \textcolor{dgreen}{$\mathcal{O}$} represents output modules, and \textcolor{dred}{$\mathcal{Q}$} represents queries in the transformer decoder.
}
\centering

\begin{tabular}{c|c|c|ccc}
\hline
                                                                                                   &                                                                                                &                             & \multicolumn{3}{c}{Application Efficiency}                                                                                      \\  
                                                                                                   &                                                                                                &                             & \multicolumn{1}{c|}{{Training}} & \multicolumn{1}{c|}{Inference}                 & Storage                 \\
\multirow{-3}{*}{Paradigms or Models}                                                              & \multirow{-3}{*}{\begin{tabular}[c]{@{}c@{}}Focus for Unified \\ Face Perception\end{tabular}} & \multirow{-3}{*}{Extensible?} & \multicolumn{1}{c|}{Cycles}                          & \multicolumn{1}{c|}{Calculation}               & Parameter               \\ \hline
                                                                                                   &                                                                                                &                             & \multicolumn{1}{c|}{}                                & \multicolumn{1}{c|}{}                          &                         \\
\multirow{-2}{*}{\begin{tabular}[c]{@{}c@{}}Universal Representation \\ + Finetuning\end{tabular}} & \multirow{-2}{*}{Representation}                                                               & \multirow{-2}{*}{Yes}       & \multicolumn{1}{c|}{\multirow{-2}{*}{$N$}}             & \multicolumn{1}{c|}{\multirow{-2}{*}{$N$\textcolor{dblue}{$\mathcal{B}$}$+N$\textcolor{dgreen}{$\mathcal{O}$}}} & \multirow{-2}{*}{$N$\textcolor{dblue}{$\mathcal{B}$}$+N$\textcolor{dgreen}{$\mathcal{O}$}} \\
Early All-In-One Model                                                                             & Training                                                                                       & No                          & \multicolumn{1}{c|}{1}                               & \multicolumn{1}{c|}{$1$\textcolor{dblue}{$\mathcal{B}$}$+N$\textcolor{dgreen}{$\mathcal{O}$}}                 & $1$\textcolor{dblue}{$\mathcal{B}$}$+N$\textcolor{dgreen}{$\mathcal{O}$}               \\
Our Naive Faceptor                                                                                 & Model Structure                                                                                & Yes                         & \multicolumn{1}{c|}{1}                               & \multicolumn{1}{c|}{$1$\textcolor{dblue}{$\mathcal{B}$}$+N$\textcolor{dgreen}{$\mathcal{O}$}}                 & $1$\textcolor{dblue}{$\mathcal{B}$}$+N$\textcolor{dgreen}{$\mathcal{O}$}               \\
Our Faceptor                                                                                       & Model Structure                                                                                & Yes                         & \multicolumn{1}{c|}{1}                               & \multicolumn{1}{c|}{$1$\textcolor{dblue}{$\mathcal{B}$}$+N$\textcolor{dgreen}{$\mathcal{O}$}}                 & $1$\textcolor{dblue}{$\mathcal{B}$}$+1$\textcolor{dgreen}{$\mathcal{O}$}$+N$\textcolor{dred}{$\mathcal{Q}$} (\textcolor{dred}{$\mathcal{Q}$}$\ll$\textcolor{dgreen}{$\mathcal{O}$})        \\ \hline
\end{tabular}
\label{table:paradigm}
\vspace{-25pt}
\end{table}

\section{Experiments}

\subsection{Implementation Details}
\label{sec:implementation}

\subsubsection{Datasets}
To validate the effectiveness of our proposed generalist models, we have collected 13 training datasets covering 6 tasks within 3 categories. 
In our experiments, Naive Faceptor and the base version of Faceptor (referred to as Faceptor-Base) are trained with only the 7 datasets highlighted in bold in \cref{table:dataset}.
To explore the performance ceiling of Faceptor, we further train Faceptor-Full using all 13 datasets.
\Cref{table:dataset} presents the number of training samples in each dataset after preprocessing.
For dense prediction, we apply the data augmentation methods used in the FaRL~\cite{FaRL}'s downstream experiment.
For attribute prediction, we employ horizontal flip, Randaugment~\cite{Randaugment}, and Random Erasing~\cite{RandomErasing}. 
For identity prediction, we use only horizontal flip for data augmentation. 
It is worth noting that we do not perform uniform alignment for training samples used but still achieve excellent performance.
Please refer to the appendix for more details of the datasets.

\begin{table}\scriptsize
    \caption{The face analysis tasks included in our experiment and the corresponding datasets used}
\centering
\begin{tabular}{c|c|c|c|cc|cc}
\hline
\multirow{2}{*}{Task Category}                                                   & \multirow{2}{*}{Task}                                                                          & \multirow{2}{*}{\begin{tabular}[c]{@{}c@{}}Datasets for \\ Training\end{tabular}} & \multirow{2}{*}{\begin{tabular}[c]{@{}c@{}}Number of \\ Samples\end{tabular}} & \multicolumn{2}{c|}{Faceptor-Base}          & \multicolumn{2}{c}{Faceptor-Full}           \\
                                                                                 &                                                                                                &                                                                                   &                                                                               & $n_t$               & $\alpha_t$            & $n_t$               & $\alpha_t$            \\ \hline
\multirow{6}{*}{\begin{tabular}[c]{@{}c@{}}Dense \\ Prediction\end{tabular}}     & \multirow{4}{*}{\begin{tabular}[c]{@{}c@{}}Landmark\\ Localization\end{tabular}}               & \textbf{300W}\cite{300W}                                                                     & 3, 148                                                                        & 4                   & 1000.00               & 4                   & 250.00                \\
                                                                                 &                                                                                                & WFLW\cite{WFLW}                                                                              & 7, 500                                                                        & -                   & -                     & 4                   & 250.00                \\
                                                                                 &                                                                                                & COFW\cite{COFW}                                                                              & 1, 345                                                                        & -                   & -                     & 4                   & 250.00                \\
                                                                                 &                                                                                                & AFLW-19\cite{AFLW-19}                                                                           & 20, 000                                                                       & -                   & -                     & 4                   & 250.00                \\ \cline{2-8} 
                                                                                 & \multirow{2}{*}{\begin{tabular}[c]{@{}c@{}}Face\\ Parsing\end{tabular}}                        & \textbf{CelebAMask-HQ}\cite{CelebAMask-HQ}                                                            & 27, 176                                                                       & 4                   & 100.00                & 4                   & 100.00                \\
                                                                                 &                                                                                                & LaPa\cite{LaPa}                                                                              & 20, 168                                                                       & -                   & -                     & 4                   & 100.00                \\ \hline
\multirow{8}{*}{\begin{tabular}[c]{@{}c@{}}Attribute \\ Prediction\end{tabular}} & \multirow{2}{*}{\begin{tabular}[c]{@{}c@{}}Age \\ Estimation\end{tabular}}                     & \textbf{MORPH II}\cite{MORPH}                                                                 & 44, 194                                                                       & 64                  & 6.00                  & 64                  & 4.00                  \\
                                                                                 &                                                                                                & UTKFace\cite{UTKFace}                                                                           & 13, 144                                                                       & -                   & -                     & 16                  & 1.00                  \\ \cline{2-8} 
                                                                                 & \multirow{3}{*}{\begin{tabular}[c]{@{}c@{}}Expression \\ Recognition\end{tabular}}             & \textbf{AffectNet}\cite{AffectNet}                                                                & 282, 829                                                                      & 64                  & 4.00                  & 64                  & 6.66                  \\
                                                                                 &                                                                                                & \textbf{RAF-DB}\cite{DLP-CNN-RAF-DB}                                                                   & 12, 271                                                                       & 16                  & 1.00                  & 16                  & 1.67                  \\
                                                                                 &                                                                                                & FERPlus\cite{FERPlus}                                                                           & 28, 127                                                                       & -                   & -                     & 16                  & 1.67                  \\ \cline{2-8} 
                                                                                 & \multirow{3}{*}{\begin{tabular}[c]{@{}c@{}}Binary \\ Attribute \\ Classification\end{tabular}} & \multirow{3}{*}{\textbf{CelebA}\cite{CelebA}}                                                  & \multirow{3}{*}{182, 637}                                                     & \multirow{3}{*}{64} & \multirow{3}{*}{2.00} & \multirow{3}{*}{64} & \multirow{3}{*}{2.00} \\
                                                                                 &                                                                                                &                                                                                   &                                                                               &                     &                       &                     &                       \\
                                                                                 &                                                                                                &                                                                                   &                                                                               &                     &                       &                     &                       \\ \hline
\begin{tabular}[c]{@{}c@{}}Identity \\ Prediction\end{tabular}                   & \begin{tabular}[c]{@{}c@{}}Face \\ Recognition\end{tabular}                                    & \textbf{MS1MV3}\cite{MS-Celeb-1M}                                                                   & 5, 179, 510                                                                   & 256                 & 5.00                  & 256                 & 5.00                  \\ \hline
\end{tabular}
\label{table:dataset}
\vspace{-10pt}
\end{table}

\subsubsection{Training for Faceptor}
For the first stage, we employ an AdamW~\cite{AdamW} optimizer for 50,000 steps, using a cosine decay learning rate scheduler and 2000 steps of linear warm-up. 
The base learning rate for the Transformer Encoder is $5.0\times10^{-5}$, and the learning rate for the remaining parts is 10 times that of the Transformer Encoder. A weight decay of 0.05 is used.
For the second stage, only 20000 steps are required, with 2000 steps reserved for linear warm-up. 
All parameters except for layer-aware embeddings are frozen. 
The other hyper-parameters remain consistent with the first stage. 
Due to the small number of parameters being trained, the second stage can be completed quickly.
\Cref{table:dataset} presents the batch size and weight used for each dataset during the training of Faceptor-Base and Faceptor-Full.
All training is conducted on 4 NVIDIA Tesla V100 GPUs.

\subsubsection{Training for Naive Faceptor}
During the training of the Naive Faceptor, we have observed that this structure is not sensitive to the weight changes of the tasks. Therefore, the weights for all tasks are set to 1.0. 
Other settings are kept consistent with the first stage of training the Faceptor-Base.

\subsection{Comparison Between Naive Faceptor and Faceptor}
\label{sec:comparison_naive}

\Cref{table:com_with_naive} presents a comparison between Naive Faceptor and Faceptor-Base in terms of parameters and performance.
Overall, Faceptor-Base demonstrates similar performance to Naive Faceptor while utilizing significantly fewer parameters.
Specifically, Faceptor exhibits slight enhancements in facial landmark localization, face parsing, age estimation, and binary attribute estimation tasks, along with a notable improvement in expression recognition by 2.80\%. 
Only for face recognition, Faceptor indicates a slight decrease.
Faceptor consists of a total of 103.2M parameters, distributed as follows: 86.8M for the transformer encoder, 14.7M for the transformer decoder, 0.5M for the pixel decoder, and 1.2M for the remaining components.
In Naive Faceptor, the standardized output heads for dense, attribute, and identity prediction tasks respectively contain approximately 39.3M, 3.4M, and 1.0M parameters. 
Consequently, Naive Faceptor encompasses a total of 178.9M parameters for the six tasks, which is 73\% more than Faceptor. 
As the number of tasks increases, this parameter difference between the two models will become even more pronounced.
The experimental results indicate that Faceptor, with higher storage efficiency and comparable performance with the naive counterpart, should be favored as a unified model structure. For this reason, we conduct larger-scale experiments in \cref{sec:performance_exp} to compare the performance of our Faceptor with specialized models.

It is worth noting that we have omitted the performance comparison of our proposed models with early all-in-one models~\cite{HyperFace,AIO}, as those early models utilized significantly simpler testing protocols that are now rarely referenced, and their task sets are also smaller. Given that our generalist models perform well on more challenging and diverse testing protocols, it is evident that our models surpass the early all-in-one models. The appendix provides further discussion on the performance of early models.

\begin{table}[!ht]\scriptsize
\vspace{-10pt}
    \caption{Comparison between Naive Faceptor and Faceptor-Base}
\centering
\begin{tabular}{l|cccccc|c}
\hline
\multirow{3}{*}{Method} & \multicolumn{3}{c|}{Landmark 300W}                                   & \multicolumn{1}{c|}{Parsing}        & \multicolumn{1}{c|}{Age}            & Expression     & Attribute                                                             \\
                        & Comm.          & Chal.          & \multicolumn{1}{c|}{Full}          & \multicolumn{1}{c|}{CelebAMask-HQ}  & \multicolumn{1}{c|}{MORPH II}       & RAF-DB         & CelebA                                                                \\
                        & \multicolumn{3}{c|}{$\mathrm{NME_{inter-ocular}}$ $\downarrow$}                                & \multicolumn{1}{c|}{F1-mean $\uparrow$}        & \multicolumn{1}{c|}{MAE $\downarrow$}            & Acc $\uparrow$            & mAcc $\uparrow$                                                                  \\ \hline
Naive                   & 2.75           & 4.84           & \multicolumn{1}{c|}{3.16}          & \multicolumn{1}{c|}{88.04}          & \multicolumn{1}{c|}{1.873}          & 87.58          & 91.40                                                                 \\
\textbf{Faceptor}       & \textbf{2.60}  & \textbf{4.60}  & \multicolumn{1}{c|}{\textbf{3.00}} & \multicolumn{1}{c|}{\textbf{88.22}} & \multicolumn{1}{c|}{\textbf{1.869}} & \textbf{90.38} & \textbf{91.43}                                                        \\ \hline
\multirow{3}{*}{Method} & \multicolumn{6}{c|}{Face Recognition}                                                                                                                             & \multirow{3}{*}{\begin{tabular}[c]{@{}c@{}}Params\\ (M)\end{tabular}} \\
                        & \multicolumn{6}{c|}{1:1 Verification Accuracy $\uparrow$}                                                                                                                    &                                                                       \\
                        & LFW            & CFP-FP         & AgeDB-30                           & CALFW                               & \multicolumn{1}{c|}{CPLFW}          & Mean           &                                                                       \\ \hline
Naive                   & 99.50          & \textbf{96.17} & \textbf{94.35}                     & \textbf{95.13}                      & \multicolumn{1}{c|}{\textbf{92.68}} & \textbf{95.57} & 178.9                                                                 \\
\textbf{Faceptor}       & \textbf{99.52} & 95.86          & 93.33                              & 94.70                               & \multicolumn{1}{c|}{92.12}          & 95.10          & \textbf{103.2} \\ \hline                                                      
\end{tabular}
\label{table:com_with_naive}
\vspace{-20pt}
\end{table}

\subsection{Layer-Attention Mechanism}
\label{sec:layer_exp}

\begin{wraptable}{r}{6cm}

\vspace{-35pt}

\begin{minipage}{0.48\textwidth}
\scriptsize
    \caption{Comparison under three settings. LA stands for Layer-Attention.}
    \centering
\resizebox{\linewidth}{!}{

\begin{tabular}{l|c|c|c}
\hline
\multirow{3}{*}{Settings}                               & Age            & Expression     & Attribute      \\
                                                      & MORPH II       & RAF-DB         & CelebA         \\
                                                      & MAE $\downarrow$            & Acc $\uparrow$            & mAcc $\uparrow$           \\ \hline
$w/o$ LA                                      & 1.882          & 89.80          & 91.40          \\
LA (Directly)          & 1.970          & 90.03          & 91.40          \\
\textbf{LA (Two-stage)} & \textbf{1.869} & \textbf{90.38} & \textbf{91.43} \\ \hline
\end{tabular}
}
\label{table:com_settings}

\end{minipage}
\vspace{-10pt}
\end{wraptable}

\Cref{table:com_settings} presents the performance of Faceptor-Base on age estimation, expression recognition, and binary attribute classification tasks under three settings: without using the Layer-Attention mechanism, using the Layer-Attention mechanism directly, and using the Layer-Attention mechanism with two-stage training process.
It can be observed that when using the Layer-Attention mechanism directly, Faceptor does not always achieve improved performance and even exhibits significant degradation in age estimation. However, employing two-stage training generally leads to performance improvement, especially in expression recognition, where a 0.58\% improvement is achieved on RAF-DB~\cite{DLP-CNN-RAF-DB}.

\subsection{Comprehensive Performance Evaluation for Faceptor}
\label{sec:performance_exp}

To explore the upper limit of Faceptor's performance, we have trained Faceptor-Full using 13 training datasets. 
\Cref{table:identity,table:attr,table:dense} present the performance of Faceptor-Full in various tasks.
In most tasks, Faceptor-Full achieves comparable or superior performance to state-of-the-art specialized models, except face recognition where it slightly lags behind the state-of-the-art method.
A detailed analysis of the performance is presented below.

\subsubsection{Dense Prediction}
Thanks to the masked image modeling ~\cite{Beit,MIM} incorporated into the FaRL framework ~\cite{FaRL}, our model achieves outstanding performance in dense prediction tasks.
Faceptor-Full outperforms existing methods on all facial landmark localization and face parsing datasets except for LaPa, as shown in \cref{table:dense}.
However, for LaPa, our model's performance declines due to the introduction of Tanh-warping~\cite{tanh} to balance segmentation performance between the inner facial components and hair region. 
We conduct experiments using Faceptor-Base for transfer learning on the LaPa dataset, achieving a mean F1 score of 92.7, as shown in \cref{table:cross}. This score is higher than that of the state-of-the-art specialized methods, demonstrating our model's strong understanding of dense prediction tasks.

\begin{table}[!ht]\mycustomsize
\vspace{-10pt}
    \caption{Comparison with other specialized models for dense prediction tasks}
    \centering

\begin{tabular}{lcccccc|lcc}
\hline
\multirow{3}{*}{Method} & WFLW          & \multicolumn{3}{c}{300W}                      & COFW          & AFLW-19       & \multirow{3}{*}{Method} & CelebA        & \multirow{2}{*}{LaPa} \\
                        & Full          & Comm.         & Chal.         & Full          & -             & Full          &                         & Mask-HQ       &                       \\
                        & \multicolumn{5}{c}{$\mathrm{NME_{inter-ocular}} \downarrow$}                                          & $\mathrm{NME_{diag}} \downarrow$      &                         & \multicolumn{2}{c}{F1-mean $\uparrow$}           \\ \hline
DAN-Menpo~\cite{DAN-Menpo}               & -             & 3.44          & 4.88          & 3.09          & -             & -             & Lee et al.~\cite{lee}              & 80.3          & -                     \\
SAN~\cite{SAN}                     & -             & 3.34          & 6.60           & 3.98          & -             & 1.91          & BASS~\cite{LaPa}                    & -             & 89.8                  \\
LAB~\cite{LAB}                     & 5.27          & 2.98          & 5.19          & 3.49          & 3.92          & 1.85          & EHANet~\cite{EHANet}                  & 84.0          & 89.2                  \\
Wing~\cite{Wing}                    & 5.11          & 3.27          & 7.18          & 4.04          & -             & 1.65          & Wei et al.~\cite{Wei}              & 82.1          & 89.4                  \\
DeCaFA~\cite{DeCaFA}                  & 4.62          & 2.93          & 5.26          & 3.39          & -             & -             & EAGR~\cite{EAGR}                    & 85.1          & 91.1                  \\
AWing~\cite{AWing}                   & 4.36          & 2.72          & 4.52          & 3.07          & -             & -             & AGRNet~\cite{AGRNet}                  & 85.5          & \underline{92.3}                  \\
AVS~\cite{AVS}+SAN~\cite{SAN}                 & 4.39          & 3.21          & 6.49          & 3.86          & -             & -             & DML-CSR~\cite{DML-CSR}                 & \underline{86.1}          & \textbf{92.4}                  \\
HRNet~\cite{HRNet}                   & 4.60          & 2.91          & 5.11          & 3.34          & 3.45          & 1.57          &                         &               &                       \\
DAG~\cite{DAG}                     & 4.21          & 2.62          & 4.77          & 3.04          & -             & -             &                         &               &                       \\
LUVLi~\cite{LUVLi}                   & 4.37          & 2.76          & 5.16          & 3.23          & -             & 1.39          &                         &               &                       \\
ADNet~\cite{ADNet}                   & 4.14          & \underline{2.53}          & 4.58          & \underline{2.93}          & -             & -             &                         &               &                       \\
PIPNet~\cite{PIPNet}                  & 4.31          & 2.78          & 4.89          & 3.19          & 3.08          & 1.42          &                         &               &                       \\
SLPT~\cite{SLPT}                    & 4.14          & 2.75          & 4.90          & 3.17          & 3.32          & -             &                         &               &                       \\
DTLD+~\cite{DTLD}                    & \underline{4.05}          & 2.60          & \underline{4.48}          & 2.96          & \underline{3.02}          & \underline{1.37}          &                         &               &                       \\ \hline
\textbf{Faceptor}       & \textbf{4.03} & \textbf{2.52} & \textbf{4.25} & \textbf{2.86} & \textbf{3.01} & \textbf{0.95} & \textbf{Faceptor}       & \textbf{88.2} & 91.5         \\ \hline
\end{tabular}
\label{table:dense}
\vspace{-20pt}
\end{table}

\begin{table}[!ht]\mycustomsize
    \caption{Comparison with other specialized models for attribute prediction tasks}
    \centering

\begin{tabular}{lcc|lcc|lc}
\hline
\multirow{3}{*}{Methods} & \multicolumn{2}{c|}{Age}      & \multirow{3}{*}{Methods} & \multicolumn{2}{c|}{Expression}              & \multirow{3}{*}{Methods} & Attribute            \\
                        & MORPH II      & UTKFace       &                         & RAF-DB               & FERPlus               &                         & CelebA               \\
                        & \multicolumn{2}{c|}{MAE $\downarrow$}      &                         & \multicolumn{2}{c|}{Acc $\uparrow$}                     &                         & mAcc $\uparrow$                 \\ \hline
OR-CNN~\cite{OR-CNN}                  & 3.27          & 5.74          & DLP-CNN~\cite{DLP-CNN-RAF-DB}                 & 80.89                & -                     & PANDA-1~\cite{PANDA-1}                 & 85.43                \\
DEX~\cite{DEX}                     & 2.68          & -             & gACNN~\cite{gACNN}                   & 85.07                & -                     & LNets+ANet~\cite{CelebA}              & 87.33                \\
DLDL~\cite{DLDL}                   & 2.42          & -             & IPA2LT~\cite{IPA2LT}                  & 86.77                & -                     & MOON~\cite{MOON}                    & 90.94                \\
DLDLF~\cite{DLDLF}                   & 2.24          & -             & RAN~\cite{RAN}                     & 86.90                & 88.55                 & NSA~\cite{NSA}                     & 90.61                \\
DRFs~\cite{DRFs}                    & 2.17          & -             & CovPool~\cite{CovPool}                 & 87.00                & -                     & MCNN-AUX~\cite{MCNN-AUX}                & 91.29                \\
MV~\cite{MV}                      & 2.16          & -             & SCN~\cite{SCN}                     & 87.03                & 89.35                 & MCFA~\cite{MCFA}                    & 91.23                \\
Axel Berg et al.~\cite{Axel}        & -             & 4.55          & DACL~\cite{DACL}                    & 87.78                & -                    & DMM-CNN~\cite{DMM-CNN}                 & \textbf{91.70}       \\
CORAL~\cite{CORAL}                   & -             & 5.47          & KTN~\cite{KTN}                     & 88.07                & \textbf{90.49}        & SwinFace~\cite{SwinFace}                & 91.32                \\
Gustafsson et al.~\cite{Gustafsson}       & -             & 4.65          & DMUE~\cite{DMUE}                    & 88.76                & 88.64                 &                         & \textbf{}            \\
BridgeNet~\cite{BridgeNet}               & 2.38          & -             & RUL~\cite{RUL}                     & 88.98                & 88.75                 &                         & \multicolumn{1}{l}{} \\
OL~\cite{OL}                      & 2.22          & -             & EAC~\cite{EAC}                     & 88.99                & 89.64                 &                         & \multicolumn{1}{l}{} \\
DRC-ORID~\cite{DRC-ORID}                & 2.16          & -             & SwinFace~\cite{SwinFace}                & {\ul 90.97}          & -                     &                         &                      \\
PML~\cite{PML}                     & 2.15          & -             &                         & \multicolumn{1}{l}{} & \multicolumn{1}{l|}{} &                         &                      \\
DLDL-v2~\cite{DLDLv2}                 & {\ul 1.97}    & 4.42          &                         & \multicolumn{1}{l}{} & \multicolumn{1}{l|}{} &                         &                      \\
MWR~\cite{MWR}                     & 2.00          & {\ul 4.37}    &                         & \multicolumn{1}{l}{} & \multicolumn{1}{l|}{} &                         &                      \\ \hline
\textbf{Faceptor}  & \textbf{1.96} & \textbf{4.10} & \textbf{Faceptor}  & \textbf{91.26}       & {\ul 90.40}           & \textbf{Faceptor}  & {\ul 91.39}          \\ \hline
\end{tabular}
\label{table:attr}
\end{table}

\subsubsection{Attribute Prediction}
Faceptor-Full achieves state-of-the-art results in age estimation and expression recognition with 1.96 and 4.10 MAE on MORPH II~\cite{MORPH} and UTKFace~\cite{UTKFace} respectively, and 91.26\% accuracy on RAF-DB~\cite{DLP-CNN-RAF-DB}, while it performs on par with the state-of-the-art on binary attribute classification.
The training samples used for age estimation and expression recognition are insufficient relative to the complexity of these tasks.
During joint training, these tasks can benefit from the initialization of universal representation and multi-task learning, obtaining improved performances.
In contrast, for the binary attribute classification task, the availability of ample data from CelebA~\cite{CelebA} with around 183K training samples has led to saturated performance across existing methods.

\begin{table}[!ht]\scriptsize
    \caption{Comparison for face recognition. The 1:1 verification accuracies on the LFW~\cite{LFW}, CFP-FP~\cite{CFP-FP}, AgeDB-30~\cite{AgeDB-30}, CALFW~\cite{CALFW} and CPLFW~\cite{CPLFW} are provided.}
    \centering
\begin{tabular}{l|cccccc}
\hline
\multirow{2}{*}{Methods} & \multicolumn{6}{c}{Face Verification Accuracy}                             \\
                        & LFW   & CFP-FP & AgeDB-30 & CALFW & \multicolumn{1}{c|}{CPLFW} & Mean \\ \hline
ViT~\cite{ViT}+CosFace~\cite{CosFace}             & 99.83 & 96.19  & 97.82    & 95.92 & \multicolumn{1}{c|}{92.55} & 96.46  \\
FaRL~\cite{FaRL}+CosFace~\cite{CosFace}            & 99.60 & 96.70  & 95.55    & 95.43 & \multicolumn{1}{c|}{92.38} & 95.93  \\
\textbf{Faceptor}                & 99.40 & 96.34  & 93.65    & 94.75 & \multicolumn{1}{c|}{92.27} & 95.28   \\ \hline
\end{tabular}
\label{table:identity}
\vspace{-10pt}
\end{table}

\subsubsection{Identity Prediction}
The performances of specialized models trained using the MS-Celeb-1M~\cite{MS-Celeb-1M} dataset and the CosFace~\cite{CosFace} loss function starting from randomly initialized ViT-B~\cite{ViT} and FaRL pretraining are presented in \cref{table:identity}, allowing a fair comparison to Faceptor-Full. 
Evaluation results on several face verification test datasets indicate that Faceptor-Full performs lower than ViT trained from scratch.
This performance decline can be attributed to two main reasons. 
Firstly, Faceptor-Full is initialized from FaRL, which provides facial representations combining high-level and low-level information not specifically tailored for the face recognition task. 
The inferior performances of specialized models starting from FaRL pre-training compared to those trained from scratch validate this point.
Secondly, Faceptor-Full involves tasks that inherently have conflicting objectives. 
While face recognition requires the model to learn to extract identity representations ignoring variations in facial texture and movements, face dense and attribute prediction tasks demand the opposite. 
Despite the slight decline in face recognition, Faceptor-Full achieves or surpasses state-of-the-art results in all other tasks,  underscoring the significant potential of the proposed face generalist model with a highly unified model structure.

\subsection{Auxiliary Supervised Learning}
\label{sec:auxiliary_exp}

The performance improvement of certain attribute prediction tasks is limited due to insufficient data, with age estimation and expression recognition being two typical tasks.
In our experiment, we consider these two tasks as the main tasks and introduce auxiliary tasks such as facial landmark localization, face parsing, and face recognition to provide additional supervised signals.
Our results (as shown in \cref{table:aux}) show that Faceptor with auxiliary supervised learning outperforms the same model which is under single-task or multi-task learning settings. 
Moreover, our model achieves significant improvements over the state-of-the-art in age and expression tasks, with an MAE of 1.787 on MORPH II~\cite{MORPH}, reducing by 0.183, and an accuracy of 91.92\% on RAF-DB~\cite{DLP-CNN-RAF-DB}, increasing by 0.95\%.
This indicates that our proposed method can effectively enhance data efficiency by leveraging rich supervised signals from auxiliary tasks, thus enabling better performance for main tasks with insufficient data.
For more experimental details on auxiliary supervised learning, please refer to the appendix.

\begin{table}[htbp]\scriptsize

    \centering
    \begin{minipage}{0.44\textwidth}
        \centering
        \caption{Comparison for auxiliary supervised learning. STL is short for Single-Task Learning. MTL is short for Multi-Task Learning. ASL is short for Auxiliary Supervised Learning.}
        \resizebox{\linewidth}{!}{
\begin{tabular}{l|c|c}
\hline
\multirow{3}{*}{Methods}                 & Age            & Expression     \\
                                        & MORPH II       & RAF-DB         \\
                                        & MAE $\downarrow$           & Acc $\uparrow$            \\ \hline
SOTA (STL)                                    & 1.970~\cite{DLDLv2}          & 90.97~\cite{SwinFace}          \\
Naive Faceptor (STL)                   & 2.070          & 91.33          \\
Faceptor (STL)               & 2.238          & 91.10          \\
Faceptor (MTL)           & 1.869          & 90.38          \\
\textbf{Faceptor (ASL)} & \textbf{1.787} & \textbf{91.92} \\ \hline
\end{tabular}

}
\label{table:aux}
        
    \end{minipage}
    \hfill
    \begin{minipage}{0.53\textwidth}
        \centering
        \caption{Cross-datasets transfer performances under different settings. EM is short for Early Methods. PT is short for Prompt Tuning. DFT is short for Decoder Finetuning. FPFT is short for Full-Parameter Finetuning.}
        \resizebox{\linewidth}{!}{
\begin{tabular}{l|c|c|c}
\hline
\multirow{3}{*}{Settings}   & Landmark & Parsing  & Attribute \\
                           & AFLW-19~\cite{AFLW-19}   & LaPa~\cite{LaPa}        & LFW-73~\cite{LFW-73}  \\
                          & $\mathrm{NME_{diag}}$ $\downarrow$    & F1-mean $\uparrow$    & mAcc $\uparrow$ \\ \hline
EM                 & 1.91~\cite{SAN}   & 89.8~\cite{LaPa}     & -         \\
PT               & 1.89   & 84.0    & 85.56     \\
DFT         & 1.06   & 89.9      & 87.81    \\
FPFT    & 0.89   & 92.7     & 87.95   \\ \hline
\end{tabular}

}
\label{table:cross}        
    \end{minipage}
    \vspace{-15pt}
\end{table}

\subsection{Cross-Datasets Transfer}
\label{sec:cross_exp}

We aim to explore the performance of Faceptor in cross-dataset transfer scenarios where subtle semantic variations exist in certain tasks, as shown in \cref{table:cross}.
We have observed that facial landmark localization datasets encompass different landmarks, face parsing datasets involve varying semantic parsing classes, and binary attribute classification datasets have different attribute labels.
Starting from Faceptor-Base, we try to transfer its capabilities to unseen datasets with novel semantics. By considering the diverse trainable parameters, we investigate three settings: training only task-specific queries (prompt tuning), training only the decoders and other output structures (output module fine-tuning), and training all parameters (full-parameter fine-tuning).
The experiments reveal that in facial landmark localization, prompt tuning results even outperform the early method~\cite{SAN}. In face parsing, the results of prompt tuning can approach the performance of the early method~\cite{LaPa}. In binary attribute classification, prompt tuning can achieve performance close to that of full-parameter fine-tuning. 
These experimental findings demonstrate the potential of prompt tuning for Faceptor.
For more experimental details, please refer to the appendix.

\section{Conclusion}

To the best of our knowledge, this is the first work that explores face generalist models.
Naive Faceptor consists of one shared backbone and 3 types of standardized output heads, obtaining improved task extensibility and increased application efficiency.
Compared to Naive Faceptor, Faceptor is more unified in structure and offers higher storage efficiency with a single-encoder dual-decoder architecture and task-specific queries for semantics.
We demonstrate the effectiveness of the proposed models on a task set including 6 tasks, achieving excellent performance.
In particular, we introduce a Layer-Attention mechanism that models the preferences of different tasks towards features from different layers, thereby enhancing performance further.
The two-stage training process ensures the effectiveness of the Layer-Attention mechanism.
Additionally, our training framework can also perform auxiliary supervised learning to improve performance on attribute prediction tasks with insufficient data.


%
%
\bibliographystyle{splncs04}
\bibliography{main}

\clearpage

\appendix

\section{Supplementary Explanations of the Method}

\subsection{Categorization for Face Analysis Tasks}

Face analysis tasks can be classified into the following three categories based on the differences in shape and granularity of their expected outputs:

\begin{enumerate}
\item \textbf{Dense prediction} involves tasks like facial landmark localization, face parsing, and depth estimation that require predictions for each pixel in an image. 
\item \textbf{Attribute prediction} includes tasks such as age estimation, expression recognition, binary attribute classification (e.g., gender classification), race classification, face forgery detection, and face anti-spoofing. The prediction outcome in these tasks is a continuous or discrete label.
\item \textbf{Identity prediction}, commonly referred to as face recognition, is a basic face perception task that represents a face identity with a vector. 
\end{enumerate}

\subsection{Naive Faceptor}

\begin{figure}[!ht]
\vspace{-10pt}
\centering
\includegraphics[height=6.0cm]{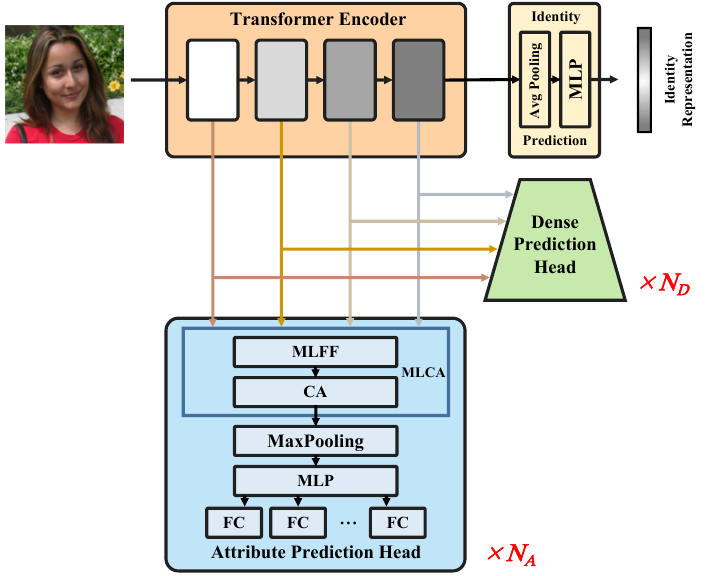}
\caption{Overall architecture for the proposed Naive Faceptor}
\label{fig:main4}
\vspace{-10pt}
\end{figure}

As shown in \cref{fig:main4}, the Naive Faceptor employs standardized face analysis and face recognition subnets from SwinFace~\cite{SwinFace} as attribute prediction head and identity prediction head, respectively.
The Multi-Level Channel Attention (MLCA) module is integrated into the attribute prediction head, which consists of a Multi-Level Feature Fusion (MLFF) module and a Channel Attention (CA) module. 
MLFF is used to combine feature maps at different levels enabling the task-specific subnet to rely on both local and global information of the faces and CA emphasizes the contributions of different levels for the specific group of tasks. 
In addition, we follow the implementation in the FaRL experiment, utilizing UperNet~\cite{UperNet} as the dense prediction head to produce dense output.

For dense prediction tasks, encoded features from the 4th, 6th, 8th, and 12th layers of the transformer encoder~\cite{ViT} are passed into dense prediction heads, while for attribute prediction tasks, features from the 6th, 8th, 10th, and 12th layers are utilized. 
In our experiments, given the involvement of two dense prediction tasks (facial landmark localization and face parsing) and three attribute prediction tasks (age estimation, expression recognition, and binary attribute classification), $N_D$ and $N_A$ are set to 2 and 3 respectively. 
In the dense prediction heads, the channel number of the output map is also an adjustable hyperparameter, configured as the number of landmarks and semantic parsing classes for the two tasks respectively.
Regarding the attribute prediction heads, the number and the output dimension of FC units are adjustable hyperparameters, set to 1 and 101 for age estimation, 1 and 7 for expression recognition, and 40 and 1 for binary attribute classification.

\subsection{Objective Functions}

In our experiments, we evaluate the effectiveness of the proposed face generalist models across a diverse set of tasks.
The objective function for each task used in our framework is as follows.

\subsubsection{Facial Landmark Localization} 
This task aims to predict heatmaps of the landmarks, as is practiced by AWing~\cite{AWing}, LUVLi~\cite{LUVLi} and ADNet~\cite{ADNet}.
We employ a loss function, combining the binary cross-entropy loss and the L1 loss: 
\begin{equation}
\scriptsize
L_{lan}=\sum_{k=1}^{N_{lan}}\{-\frac{1}{M}\sum_{m=1}^{M}[(1-p_{k,m})\log(1-\hat{p}_{k,m})+p_{k,m}\log\hat{p}_{k,m}]+\lambda[|\hat{l}_{k,x}-l_{k,x}|+|\hat{l}_{k,y}-l_{k,y}|]\},
\label{eq:landmark-loss}
\end{equation}
where $p_{k,m}=1$ if pixel $m$ of the input image is within the circle centered at landmark $k$ with a radius of 5, otherwise 0.
The output heatmap for facial landmark localization is $\mathbf{y}_{lan}\in\mathbb{R}^{N_{lan}\times H\times W}$.
The predicted probability $\hat{p}_{k,m}$ for pixel $m$ at the channel $k$ of the output heatmap is calculated with the sigmoid function. 
$M$ represents the total number of pixels, which is equal to the product of $H$ and $W$.
$(\hat{l}_{k,x},\hat{l}_{k,y})$ is the predicted result for the coordinate of landmark $k$, calculated by normalizing the channel $k$ of the output heatmap.
$(l_{k,x},l_{k,y})$ represents the ground-truth for the coordinate of landmark $k$.
$\lambda$ is a weight that balances the importance between two types of losses and is set to 1.0.

\subsubsection{Face Paring} 
This task is trained with cross-entropy loss for each pixel:
\begin{equation}
L_{par}=-\frac{1}{M}\sum_{m=1}^{M}\sum_{i=1}^{N_{par}}p_{i,m}\log\hat{p}_{i,m},
\end{equation}
where $p_{i,m}=1$ if pixel $m$ of the input image belongs to semantic parsing class $i$, otherwise 0.
The predicted probability  $\hat{p}_{i,m}$ is calculated with softmax for each pixel from the output for face parsing $\mathbf{y}_{par}\in\mathbb{R}^{N_{par}\times H\times W}$. 
$M$ represents the total number of pixels, which is equal to the product of $H$ and $W$.

\subsubsection{Age Estimation}
We train the task of age estimation by jointly learning label distribution and expectation regression, following DLDL-v2~\cite{DLDLv2}:
\begin{equation}
L_{age}=-\sum_{i=0}^{100}p_i\log\hat{p}_i+\lambda |\hat{a}-a|,
\label{eq:age-loss}
\end{equation}
where $\mathbf{p}$ is age label distribution which can be estimated with training samples.
$\hat{\mathbf{p}}$ is predicted distribution which should be similar to $\mathbf{p}$. 
We use a softmax function to turn the output for age estimation $\mathbf{y}_i\in\mathbb{R}^{101}$, into a 
predicted probability, that is, $\hat{p}_i=\frac{\exp(\mathbf{y}_{age,i})}{\sum_{j=0}^{100}\exp(\mathbf{y}_{age,j})}$.
$a_i$ is the ground-truth age.
The predicted age $\hat{a}$ can be calculated by: $\hat{a}=\sum_{i=0}^{100}i\hat{p}_i$.
$\lambda$ is a weight that balances the importance between two types of losses and is set to 1.0.

\subsubsection{Expression Recognition} 
The expressions include surprise, fear, disgust, happiness, sadness, anger, and neutral.
The loss function for training is as follows:
\begin{equation}
L_{exp}=-\sum_{i=1}^{7}p_i\log\hat{p}_i,
	\label{eq:expression-loss}
\end{equation}
where $p_i = 1$ if the input sample belongs to expression class $i$, otherwise 0.
The predicted probability is given by $\hat{p}_{i}$ which is calculated from the output for expression recognition $\mathbf{y}_{exp}\in\mathbb{R}^{7}$ by softmax.

\subsubsection{Binary Attribute Classfication} 
This task involves $N_{att}$ binary labels, and the total loss function is the sum of $N_{att}$ binary cross-entropy loss functions:
\begin{equation}
L_{att}=-\sum_{k=1}^{N_{att}}[(1-p_{k})\log(1-\hat{p}_{k})+p_{k}\log\hat{p}_{k}],
	\label{eq:attribute-loss}
\end{equation}
where $p_k = 1$ for the $k$-th attribute exists and 0 otherwise.
$p_k$ is the predicted probability that the input face contains the $k$-th attribute.
It is calculated from the $\mathbf{y}_{att}\in\mathbb{R}^{N_{att}}$ by sigmoid function.

\subsubsection{Face Recognition}
We train the task of face recognition with CosFace~\cite{CosFace}:
	\begin{equation}
		L_{rec}=-\log\frac{e^{s(\cos\theta_{i}-m)}}  {e^{s(\cos \theta_{i}-m)}+ {\textstyle \sum_{j=1,j\neq i}^{n}e^{s \cos \theta _j}}}.
		\label{eq:recognition-loss}
	\end{equation}
Initially, we feed the facial identity representation obtained from Naive Faceptor or Faceptor into a fully connected layer to predict the identity label of the sample.
The weight of the fully connected layer can be written as $\mathbf{W} \in \mathbb{R}^{d\times  n}$, where $n$ is the number of identities.
We use $\mathbf{W}_j\in \mathbb{R}^d$ to denote the $j$-th column of the weight $\mathbf{W}$
and $\mathbf{y}_{rec} \in \mathbb{R}^d$ to denote the deep feature for the input sample, belonging to the $i$-th class.
$\theta_j$ is the angle between the weight $\mathbf{W}_j$ and the feature $\mathbf{y}_{rec}$.
The embedding feature $\left \| \mathbf{y}_{rec} \right \|$  is fixed by $l_2$ normalization and re-scaled to $s$.
$m$ is the CosFace margin penalty.
In our implementation, s is set to 64, and m is set to 0.4.

\section{Implementation details}

\subsection{Datasets}

\subsubsection{300W~\cite{300W}:}
It is the most commonly used dataset for facial landmark localization which includes 3,148 images for training and 689 images for testing. The training set consists of the full set of AFW~\cite{AFW}, the training subset of HELEN~\cite{HELEN}, and LFPW~\cite{LFPW}. The test set is further divided into a challenging subset that includes 135 images (IBUG full set~\cite{300W}) and a common subset that consists of 554 images (test subset of HELEN and LFPW). Each image in 300W is annotated with 68 facial landmarks. 

\subsubsection{WFLW~\cite{WFLW}:}
It is collected from the WIDER Face dataset, encompassing large variations in pose, expression, and occlusion. It provides 98 manually annotated landmarks for 10,000 images, 7,500 for training, and 2,500 for testing. The test set is further divided into 6 subsets for different scenarios.

\subsubsection{COFW~\cite{COFW}:}
It contains 1,345 training images and 507 testing images with 29 landmarks.

\subsubsection{AFLW-19~\cite{AFLW-19}:}
The original AFLW~\cite{AFLW} provides at most 21 landmarks for each face but excludes coordinates for invisible landmarks. 
AFLW-19 provides manually annotated coordinates for these invisible landmarks. 
The new annotation does not include two ear points because it is very difficult to decide the location of invisible ears. 
This causes the point number of AFLW-19 to be 19.
The original AFLW does not provide a train-test partition. AFLW-19 adopts a partition with 20,000 images for training and 4,386 images for testing (AFLW-Full). In addition, a frontal subset (AFLW-Frontal) is proposed where all landmarks are visible (a total of 1,165 images).

\subsubsection{CelebAMask-HQ~\cite{CelebAMask-HQ}:}
CelebAMask-HQ consists of 30,000 high-resolution face images selected from the CelebA dataset. The masks of CelebAMask-HQ are manually annotated with the size of $512 \times 512$ and 19 classes.

\subsubsection{LaPa~\cite{LaPa}:}
It consists of more than 22,000 facial images with abundant variations in expression, pose, and occlusion.
Each image of LaPa is provided with an 11-category pixel-level label map and 106-point landmarks.

\subsubsection{MORPH II~\cite{MORPH}:}
It is an age estimation dataset, which contains 55,134 facial images of 13,617 subjects ranging from 16 to 77 years old. 
The entire dataset is randomly divided into five folds, with four folds allocated for training and one fold reserved for testing.

\subsubsection{UTKFace~\cite{UTKFace}:}
It provides about 20,000 facial images ranging from 0 to 116 years old. For a fair comparison, we employ the evaluation protocol using a subset of UTKFace covering faces between 21 and 60 years old as in MWR~\cite{MWR} - 13,147 for training, and 3,287 for testing. 

\subsubsection{AffectNet~\cite{AffectNet}:}
It stands as the largest publicly available dataset for facial expression recognition, comprising about 420K images with manually annotated labels. Due to significant label noise in this dataset and highly imbalanced training data distribution, we opt not to conduct testing on it. We preprocess the dataset according to the methods outlined in HSEmotion~\cite{HSEmotion} for joint training.
In our experiments, we utilize the version that includes seven classes of facial expressions.

\subsubsection{RAF-DB~\cite{DLP-CNN-RAF-DB}:}
It is a real-world expression dataset comprising 29,672 real-world facial images collected through Flickr's image search API and independently labeled by approximately 40 trained human annotators. For our experiments, we utilize the single-label subset, which consists of 15,339 expression images with six basic emotions (happiness, surprise, sadness, anger, disgust, and fear), along with the neutral expression. Among these, 12,271 images are used for training purposes, while the remaining images are reserved for testing.

\subsubsection{FERPlus~\cite{FERPlus}:}
FERPlus~\cite{FERPlus} is extended from FER2013~\cite{FER2013} which is a large-scale dataset collected by APIs in the Google image 
search. It contains 28,709 training, 3,589 validation, and 3,589 test images. 
In our experiments, we utilize the version that includes seven classes of facial expressions.

\subsubsection{CelebA~\cite{CelebA}:}
It is a large-scale collection of facial attributes, comprising 162,770 images for training, 19,867 images for validation, and 19,962 images for testing. Each image in CelebA is extensively annotated with 40 binary attributes.

\subsubsection{LFW-73~\cite{LFW-73}:}
It is another challenging facial dataset, comprising 13,143 images annotated with 73 binary facial attributes, 40 of which are shared with CelebA. This dataset is divided in half for training (6,263 images) and testing (6,880 images). We utilize this dataset in the cross-datasets transfer experiment.

\subsubsection{MS-Celeb-1M~\cite{MS-Celeb-1M}:} 
It is one of the most popular training datasets in the field of facial recognition and we utilize the clean version refined by insightface~\cite{ArcFace}, containing 5.3M images of 93,431 celebrities.

\subsubsection{Face Verification Datasets:}
LFW~\cite{LFW} database contains 13,233 face images from 5,749 different identities, which is a classic benchmark for unconstrained face verification. CFP-FP~\cite{CFP-FP} and CPLFW~\cite{CPLFW} are built to emphasize the cross-pose challenge while AgeDB-30~\cite{AgeDB-30} and CALFW~\cite{CALFW} are built for the cross-age challenge. 

\subsection{Auxiliary Supervised Learning}

In the setting of auxiliary supervised learning, we consider age estimation and expression recognition as the main tasks respectively, while facial landmark localization, face parsing, and face recognition serve as auxiliary tasks.
The batch size and weight used for each dataset is presented in \cref{table:aux_dataset}.
Other hyper-parameters are kept consistent with the first stage of training the Faceptor-Base.

\begin{table}\scriptsize
\vspace{-10pt}
    \caption{The batch size and weight used for each dataset in the setting of auxiliary supervised learning}
\centering

\begin{tabular}{c|c|c|cccc}
\hline
\multirow{3}{*}{Category}        & \multirow{3}{*}{Task}                   & \multirow{3}{*}{Dataset} & \multicolumn{4}{c}{Main Task}                                                        \\ \cline{4-7} 
                                 &                                         &                          & \multicolumn{2}{c|}{Age}     & \multicolumn{2}{c}{Expression} \\
                                 &                                         &                          & $n_t$ & \multicolumn{1}{c|}{$\alpha_t$} & $n_t$             & $\alpha_t$             \\ \hline
\multirow{3}{*}{Main Task}       & Age Estimation                          & MORPH II~\cite{MORPH}                 & 64    & \multicolumn{1}{c|}{10.0}       & -                 & -                      \\ \cline{2-7} 
                                 & \multirow{2}{*}{Expression Recognition} & AffectNet~\cite{AffectNet}                & -     & \multicolumn{1}{c|}{-}          & 64                & 80.0                   \\
                                 &                                         & RAF-DB~\cite{DLP-CNN-RAF-DB}                   & -     & \multicolumn{1}{c|}{-}          & 16                & 20.0                   \\ \hline
\multirow{3}{*}{Auxiliary Tasks} & Landmark Localization                   & 300W~\cite{300W}                     & 4     & \multicolumn{1}{c|}{1000.0}     & 2                 & 1000.0                 \\ \cline{2-7} 
                                 & Face Parsing                            & CelebAMask-HQ~\cite{CelebAMask-HQ}            & 4     & \multicolumn{1}{c|}{100.0}      & 2                 & 100.0                  \\ \cline{2-7} 
                                 & Face Recognition                        & MS1MV3~\cite{MS-Celeb-1M}                   & 256   & \multicolumn{1}{c|}{10.0}       & 64                & 1.0                    \\ \hline
\end{tabular}

\label{table:aux_dataset}
\vspace{-20pt}
\end{table}

\subsection{Cross-Datasets Transfer}
Starting from Faceptor-Base, cross-dataset transfer experiments are conducted on AFLW-19~\cite{AFLW-19}, LaPa~\cite{LaPa}, and LFW-73~\cite{LFW-73} with batch sizes set to 8, 8, and 32 respectively.
20000 steps are required for tuning, with 2000 steps reserved for linear warm-up.
Other hyper-parameters are kept consistent with the first stage of training the Faceptor-Base.

\section{Additional Results}

\subsection{Performance of Early All-In-One Models}

Early all-in-one models~\cite{HyperFace,AIO} employ significantly simpler testing protocols that are now rarely referenced. 
In this section, we provide a detailed discussion of the performance of these early models on the tasks they can address.
Through indirect comparison, we have demonstrated that the proposed Faceptor outperforms these early all-in-one models significantly.

\subsubsection{Facial Landmark Localization}
HyperFace~\cite{HyperFace} and AIO~\cite{AIO} report performance for facial landmark localization on AFW~\cite{AFW} and the original AFLW~\cite{AFLW} datasets. 
AFW contains only 205 images with 468 faces. 
The full set of AFW has already been incorporated into the training samples of the 300W~\cite{300W} protocol. 
By manually annotating coordinates for invisible landmarks, the original AFLW dataset has been reprocessed into the more commonly used testing protocol known as AFLW-19~\cite{AFLW-19}.
Our Faceptor has achieved performance surpassing the state-of-the-art method on more challenging 300W and AFLW-19.
Although results on AFW and the original AFLW dataset are not reported, it is evident that our Faceptor significantly outperforms early all-in-one methods in facial landmark localization.

\subsubsection{Age Estimation}

AIO~\cite{AIO} provides test results on CLAP2015~\cite{CLAP2015} and FG-NET~\cite{FG-Net} datasets. 
CLAP2015 consists of 2,476 training samples and 1,079 testing samples. 
FG-NET contains a total of 1,002 face samples and is commonly used for leave-one-person-out testing protocol. 
To ensure an adequate number of training and testing examples, we employ the MORPH II~\cite{MORPH} and UTKFace~\cite{UTKFace} protocols to evaluate the performance of our proposed models in age estimation. 
By providing results of MWR~\cite{MWR} on MORPH II, UTKFace, and CLAP2015, we indirectly demonstrate the superior age estimation capabilities of our Faceptor compared to the early all-in-one model.

\begin{table}[]\scriptsize
\vspace{-10pt}
    \caption{Comparison for age estimation}
\centering
\begin{tabular}{l|cc|c}
\hline
\multirow{2}{*}{Methods} & MORPH II      & UTKFace       & CLAP2015      \\
                         & \multicolumn{2}{c|}{MAE $\downarrow$}       & $\epsilon$-error $\downarrow$       \\ \hline
AIO~\cite{AIO}                      & -             & -             & 0.29          \\
MWR~\cite{MWR}                      & 2.00          & 4.37          & \textbf{0.26} \\
\textbf{Faceptor-Full}   & \textbf{1.96} & \textbf{4.10} & -             \\ \hline
\end{tabular}

\vspace{-20pt}
\end{table}

\subsubsection{Binary Attribute Classification}
CelebA~\cite{CelebA} is the most commonly used binary attribute classification dataset. HyperFace~\cite{HyperFace} supports only gender classification, while AIO~\cite{AIO} supports gender and smile classification. Our Faceptor supports all 40 attribute classification tasks involved in CelebA. Even in gender and smile classification tasks, our method achieves the same accuracy as AIO.

\begin{table}[]\scriptsize
\vspace{-10pt}
    \caption{Comparison for binary attribute classification}
\centering
\begin{tabular}{l|cc|c}
\hline
\multirow{2}{*}{Methods} & Gender      & Smile       & All Attributes \\
                         & \multicolumn{2}{c|}{Acc $\uparrow$}  & mAcc $\uparrow$           \\ \hline
HyperFace~\cite{HyperFace}                & 97          & -           & -              \\
AIO~\cite{AIO}                      & \textbf{99} & \textbf{93} & -              \\
\textbf{Faceptor-Full}   & \textbf{99} & \textbf{93} & 91.39          \\ \hline
\end{tabular}
\vspace{-10pt}
\end{table}

\subsubsection{Face Recognition}
AIO~\cite{AIO} evaluates face recognition on IJB-A~\cite{IJB-A}. 
The dataset has been extended to IJB-C~\cite{IJB-C}, which is more challenging.
By providing the results of VGGFace2~\cite{VGGFace2} on IJB-A and IJB-C, we indirectly demonstrate the superior face recognition capability of our method compared to the early all-in-one model.

\begin{table}[]\scriptsize
\vspace{-10pt}
    \caption{Comparison for face recognition}
\centering

\begin{tabular}{l|cc|ccc}
\hline
\multirow{2}{*}{Methods} & \multicolumn{2}{c|}{IJB-C TAR@FAR $\uparrow$} & \multicolumn{3}{c}{IJB-A TAR@FAR $\uparrow$} \\
                         & 0.001                   & 0.01                    & 0.001          & 0.01           & 0.1            \\ \hline
AIO~\cite{AIO}                      & -                       & -                       & 78.7           & 89.3           & 96.8           \\
VGGFace2~\cite{VGGFace2}                  & 92.7                    & 96.7                    & \textbf{92.1}  & \textbf{96.8}  & \textbf{99.0}  \\
\textbf{Faceptor-Full}   & \textbf{95.7}           & \textbf{98.1}           & -              & -              & -              \\ \hline
\end{tabular}

\vspace{-20pt}
\end{table}

\subsection{Performance Evaluation for Faceptor}

Due to space limitations, we do not include the complete test results of the Faceptor-Full on dense prediction task in the main body of the paper. Here, \cref{table:wflw_300w,table:cofw_aflw,table:celebam,table:lapa} present the complete results on datasets WFLW~\cite{WFLW}, 300W~\cite{300W}, COFW~\cite{COFW}, AFLW-19~\cite{AFLW-19}, CelebAMask-
HQ~\cite{CelebAMask-HQ} and Lapa~\cite{LaPa}.

\begin{table}[!ht]\scriptsize
\vspace{-10pt}
    \caption{Comparison with other specialized facial landmark localization methods on WFLW~\cite{WFLW} and 300W~\cite{300W}}
\centering

\begin{tabular}{l|ccccccccc|ccc}
\hline
\multirow{3}{*}{Methods} & \multicolumn{9}{c|}{WFLW}                                                                                                                                           & \multicolumn{3}{c}{300W}                      \\
                        & \multicolumn{7}{c|}{$\mathrm{NME_{inter-ocular}}$ $\downarrow$}                                                                                              & FR$^{10}$ $\downarrow$          & AUC$^{10}$ $\uparrow$          & \multicolumn{3}{c}{$\mathrm{NME_{inter-ocular}}$ $\downarrow$}          \\
                        & Full          & Pose          & Expr.         & Illum.        & M.U.        & Occl.         & \multicolumn{1}{c|}{Blur}          & \multicolumn{2}{c|}{Full}      & Comm.         & Chal.         & Full          \\ \hline
DAN-Menpo~\cite{DAN-Menpo}               & -             & -             & -             & -             & -             & -             & \multicolumn{1}{c|}{-}             & -             & -              & 3.44          & 4.88          & 3.09          \\
SAN~\cite{SAN}                     & -             & -             & -             & -             & -             & -             & \multicolumn{1}{c|}{-}             & -             & -              & 3.34          & 6.60          & 3.98          \\
LAB~\cite{LAB}                     & 5.27          & 10.24         & 5.51          & 5.23          & 5.15          & 6.79          & \multicolumn{1}{c|}{6.32}          & 7.56          & 53.23          & 2.98          & 5.19          & 3.49          \\
Wing~\cite{Wing}                    & 5.11          & 8.75          & 5.36          & 4.93          & 5.41          & 6.37          & \multicolumn{1}{c|}{5.81}          & 6.00          & 55.40          & 3.27          & 7.18          & 4.04          \\
DeCaFA~\cite{DeCaFA}                  & 4.62          & -             & -             & -             & -             & -             & \multicolumn{1}{c|}{-}             & 4.84          & 56.30          & 2.93          & 5.26          & 3.39          \\
Awing~\cite{AWing}                   & 4.36          & 7.38          & 4.58          & 4.32          & 4.27          & 5.19          & \multicolumn{1}{c|}{4.96}          & 2.84          & 57.19          & 2.72          & 4.52          & 3.07          \\
AVS~\cite{AVS}+SAN~\cite{SAN}                 & 4.39          & 8.42          & 4.68          & 4.24          & 4.37          & 5.60          & \multicolumn{1}{c|}{4.86}          & 4.08          & 59.13          & 3.21          & 6.49          & 3.86          \\
HRNet~\cite{HRNet}                   & 4.60          & 7.86          & 4.78          & 4.57          & 4.26          & 5.42          & \multicolumn{1}{c|}{5.36}          & -             & -              & 2.91          & 5.11          & 3.34          \\
DAG~\cite{DAG}                     & 4.21          & 7.36          & 4.49          & 4.12          & 4.05          & 4.98          & \multicolumn{1}{c|}{4.82}          & 3.04          & 58.93          & 2.62          & 4.77          & 3.04          \\
LUVLi~\cite{LUVLi}                   & 4.37          & -             & -             & -             & -             & -             & \multicolumn{1}{c|}{-}             & 3.12          & 57.70          & 2.76          & 5.16          & 3.23          \\
ADNet~\cite{ADNet}                   & 4.14          & 6.96          & 4.38          & 4.09          & 4.05          & 5.06          & \multicolumn{1}{c|}{4.79}          & 2.72          & 60.22          & 2.53          & 4.58          & 2.93          \\
PIPNet~\cite{PIPNet}                  & 4.31          & 7.51          & 4.44          & 4.19          & 4.02          & 5.36          & \multicolumn{1}{c|}{5.02}          & -             & -              & 2.78          & 4.89          & 3.19          \\
SLPT~\cite{SLPT}                    & 4.14          & -             & -             & -             & -             & -             & \multicolumn{1}{c|}{-}             & 2.76          & 59.50          & 2.75          & 4.90          & 3.17          \\
DTLD+~\cite{DTLD}                   & 4.05          & -             & -             & -             & -             & -             & \multicolumn{1}{c|}{-}             & 2.68          & -              & 2.60          & 4.48          & 2.96          \\ \hline
\textbf{Faceptor-Full}       & \textbf{4.03} & \textbf{6.81} & \textbf{4.28} & \textbf{3.93} & \textbf{3.91} & \textbf{4.71} & \multicolumn{1}{c|}{\textbf{4.56}} & \textbf{1.92} & \textbf{60.24} & \textbf{2.52} & \textbf{4.25} & \textbf{2.86} \\ \hline
\end{tabular}
\label{table:wflw_300w}
\vspace{-20pt}
\end{table}

\begin{table}[!ht]\scriptsize

\vspace{-10pt}
    \caption{Comparison with other specialized facial landmark localization methods on COFW~\cite{COFW} and AFLW-19~\cite{AFLW-19}}
\centering

\begin{tabular}{l|c|cccc}
\hline
\multirow{3}{*}{Methods} & COFW                              & \multicolumn{4}{c}{AFLW-19}                                                         \\
                        & \multirow{2}{*}{$\mathrm{NME_{inter-ocular}}$ $\downarrow$} & \multicolumn{2}{c|}{$\mathrm{NME_{diag}}$ $\downarrow$}                      & $\mathrm{NME_{box}}$ $\downarrow$       & $\mathrm{AUC_{box}}$ $\uparrow$        \\
                        &                                   & Full          & \multicolumn{1}{c|}{Frontal}       & \multicolumn{2}{c}{Full}       \\ \hline
SAN~\cite{SAN}                     & -                                 & 1.91          & \multicolumn{1}{c|}{1.85}          & 4.04          & 54.00          \\
LAB~\cite{LAB}                     & 3.92                              & 1.85          & \multicolumn{1}{c|}{1.62}          & -             & -              \\
Wing~\cite{Wing}                    & -                                 & 1.65          & \multicolumn{1}{c|}{-}             & -             & -              \\
HRNet~\cite{HRNet}                   & 3.45                              & 1.57          & \multicolumn{1}{c|}{1.46}          & -             & -              \\
LUVLi~\cite{LUVLi}                   & -                                 & 1.39          & \multicolumn{1}{c|}{1.19}          & 2.28          & 68.00          \\
PIPNet~\cite{PIPNet}                  & 3.08                              & 1.42          & \multicolumn{1}{c|}{-}             & -             & -              \\
SLPT~\cite{SLPT}                    & 3.32                              & -             & \multicolumn{1}{c|}{-}             & -             & -              \\
DTLD+~\cite{DTLD}                   & 3.02                              & 1.37          & \multicolumn{1}{c|}{-}             & -             & -              \\ \hline
\textbf{Faceptor-Full}       & \textbf{3.01}                     & \textbf{0.95} & \multicolumn{1}{c|}{\textbf{0.87}} & \textbf{1.35} & \textbf{81.11} \\ \hline
\end{tabular}
\label{table:cofw_aflw}

\vspace{-20pt}
\end{table}

\begin{table}[!ht]\scriptsize

    \caption{Comparison with other specialized face parsing methods on CelebAMask-HQ~\cite{CelebAMask-HQ}. Results are reported in F1 scores (\%)}
    \centering
\begin{tabular}{l|ccccccccc|c}
\hline
\multirow{2}{*}{Methods}            & Face          & Nose          & Classes       & L-Eye         & R-Eye         & L-B           & R-B           & L-Ear         & R-Ear         & \multirow{2}{*}{Mean}          \\
                                   & I-M           & U-L           & L-L           & Hair          & Hat           & Earring       & Necklace      & Neck          & Cloth         &                                \\ \hline
\multirow{2}{*}{EHANet~\cite{EHANet}}            & 96.0          & 93.7          & 90.6          & 86.2          & 86.5          & 83.2          & 83.1          & 86.5          & 84.1          & \multirow{2}{*}{84.0}          \\
                                   & {\ul 93.8}    & 88.6          & 90.3          & 93.9          & 85.9          & 67.8          & 30.1          & 88.8          & 83.5          &                                \\ \hline
\multirow{2}{*}{Wei et al.~\cite{Wei}}        & 96.4          & 91.9          & 89.5          & 87.1          & 85.0          & 80.8          & 82.5          & 84.1          & 83.3          & \multirow{2}{*}{82.1}          \\
                                   & 90.6          & 87.9          & 91.0          & 91.1          & 83.9          & 65.4          & 17.8          & 88.1          & 80.6          &                                \\ \hline
\multirow{2}{*}{EAGR~\cite{EAGR}}              & 96.2          & \textbf{94.0} & 92.3          & 88.6          & 88.7          & {\ul 85.7}    & 85.2          & 88.0          & 85.7          & \multirow{2}{*}{85.1}          \\
                                   & \textbf{95.0} & 88.9          & \textbf{91.2} & 94.9          & 87.6          & 68.3          & 27.6          & 89.4          & 85.3          &                                \\ \hline
\multirow{2}{*}{AGRNet~\cite{AGRNet}}            & {\ul 96.5}    & {\ul 93.9}    & 91.8          & {\ul 88.7}    & {\ul 89.1}    & 85.5          & 85.6          & 88.1          & {\ul 88.7}    & \multirow{2}{*}{85.5}          \\
                                   & 92.0          & \textbf{89.1} & {\ul 91.1}    & {\ul 95.2}    & 87.2          & 69.6          & 32.8          & 89.9          & 84.9          &                                \\ \hline
\multirow{2}{*}{DML-CSR~\cite{DML-CSR}}           & 95.7          & {\ul 93.9}    & 92.6          & \textbf{89.4} & \textbf{89.6} & 85.5          & {\ul 85.7}    & {\ul 88.3}    & 88.2          & \multirow{2}{*}{{\ul 86.1}}    \\
                                   & 91.8          & 87.4          & 91.0          & 94.5          & 88.5          & {\ul 71.4}    & {\ul 40.6}    & 89.6          & {\ul 85.7}    &                                \\ \hline
\multirow{2}{*}{\textbf{Faceptor-Full}} & \textbf{96.6} & {\ul 93.9}    & \textbf{94.0} & \textbf{89.4} & {\ul 89.1}    & \textbf{86.2} & \textbf{86.3} & \textbf{88.4} & \textbf{88.8} & \multirow{2}{*}{\textbf{88.2}} \\
                                   & 91.6          & {\ul 89.0}    & 90.6          & \textbf{96.2} & {\ul 90.8}    & \textbf{72.5} & \textbf{61.6} & {\ul 92.4}    & \textbf{91.0} &                                \\ \hline
\end{tabular}
\label{table:celebam}
\end{table}

\begin{table}[!ht] \scriptsize
    \caption{Comparison with other specialized face parsing methods on LaPa~\cite{LaPa}. Results are reported in F1 scores (\%)}
\centering
\begin{tabular}{l|cccccccccc|c}
\hline
Methods                           & Skin          & Hair          & L-E           & R-E           & U-L           & I-M           & L-L           & Nose          & L-B           & R-B           & Mean          \\ \hline
BASS~\cite{LaPa}                              & 97.2          & 96.3          & 88.1          & 88.0          & 84.4          & 87.6          & 85.7          & 95.5          & 87.7          & 87.6          & 89.8          \\
EHANet~\cite{EHANet}                            & 95.8          & 94.3          & 87.0          & 89.1          & 85.3          & 85.6          & 88.8          & 94.3          & 85.9          & 86.1          & 89.2          \\
Wei et al.~\cite{Wei}                        & 96.1          & 95.1          & 88.9          & 87.5          & 83.1          & 89.2          & 83.8          & 96.1          & 86.0          & 87.8          & 89.4          \\
EAGR~\cite{EAGR}                              & 97.3          & 96.2          & 89.5          & 90.0          & {\ul 88.1}    & 90.0          & 89.0          & 97.1          & 86.5          & 87.0          & 91.1          \\
AGRNet~\cite{AGRNet}                            & \textbf{97.7} & \textbf{96.5} & 91.6          & 91.1          & \textbf{88.5} & \textbf{90.7} & {\ul 90.1}    & {\ul 97.3}    & 89.9          & 90.0          & 92.3          \\
DML-CSR~\cite{DML-CSR}                           & {\ul 97.6}    & {\ul 96.4}    & {\ul 91.8}    & {\ul 91.5}    & 88.0          & 90.5          & 89.9          & {\ul 97.3}    & {\ul 90.4}    & {\ul 90.4}    & {\ul 92.4}    \\ \hline
\textbf{Faceptor-Full}            & 97.5          & 96.0          & 91.2          & 90.8          & 86.7          & 88.9          & 89.3          & 97.0          & 89.0          & 88.9          & 91.5          \\
\textbf{Faceptor-Base+Finetuning} & \textbf{97.7} & 96.3          & \textbf{92.3} & \textbf{92.2} & \textbf{88.5} & {\ul 90.6}    & \textbf{90.5} & \textbf{97.5} & \textbf{90.7} & \textbf{90.5} & \textbf{92.7} \\ \hline
\end{tabular}
\label{table:lapa}
\end{table}

\end{document}